\documentclass{article}

\usepackage{arxiv}

\usepackage[utf8]{inputenc} 
\usepackage[T1]{fontenc}    
\usepackage{hyperref}       
\usepackage{url}            
\usepackage{booktabs}       
\usepackage{amsfonts}       
\usepackage{nicefrac}       
\usepackage{microtype}      
\usepackage{lipsum}		

\usepackage{graphicx}
\usepackage{natbib}
\usepackage{doi}
\usepackage{amssymb}
\usepackage{url,hyperref,lineno,microtype,subcaption}
\usepackage[onehalfspacing]{setspace}
\usepackage{graphicx}
\usepackage{CJKutf8}
\usepackage{amsmath}
\usepackage{dirtytalk}
\usepackage{algorithm}
\usepackage{varwidth}
\usepackage{mathtools}
\usepackage{siunitx}
\usepackage[utf8]{inputenc}
\usepackage{tabto}
\usepackage{algpseudocode} 
\newcounter{algsubstate}
\renewcommand{\thealgsubstate}{\alph{algsubstate}}

\DeclareMathOperator*{\argmaxA}{argmax}

\title{Morality-based Assertion and Homophily on Social Media: A Cultural Comparison between English and Japanese Languages}

\author{Maneet Singh \\
	Department of Computer Science and Engineering \\
	Indian Institute of Technology, Ropar, India \\
	\texttt{2018csz0008@iitrpr.ac.in} \\
	\And Rishemjit Kaur\\
	CSIR-Central Scientific Instruments Organisation, Chandigarh, India \\
	Academy of Scientific and Innovative Research, Ghaziabad, India \\
	\texttt{rishemjit.kaur@csio.res.in} \\
	\And
	Akiko Matsuo \\
	Department of Psychology \\
	Tokai Gakuen University, Nagoya, Japan \\
	\texttt{matsuo-a@tokaigakuen-u.ac.jp} \\
	\And
	S.R.S. Iyengar\\
	Department of Computer Science and Engineering \\
	Indian Institute of Technology, Ropar, India \\
	\texttt{sudarshan@iitrpr.ac.in} \\
	\And
	Kazutoshi Sasahara\\
	Department of Innovation Science, School of Environment and Society\\
	Tokyo Institute of Technology, Tokyo, Japan \\
	\texttt{sasahara.k.aa@m.titech.ac.jp} \\
}

\date{}

\renewcommand{\headeright}{}
\renewcommand{\undertitle}{}
\renewcommand{\shorttitle}{\textit{Morality-based Assertion and Homophily on Social Media} }

\hypersetup{
pdftitle={A template for the arxiv style},
pdfsubject={q-bio.NC, q-bio.QM},
pdfauthor={David S.~Hippocampus, Elias D.~Striatum},
pdfkeywords={First keyword, Second keyword, More},
}

\begin{document}
\maketitle

\begin{abstract}
Moral psychology is a domain that deals with moral identity, appraisals and emotions. Previous work has primarily focused on moral development and the associated role of culture. Knowing that language is an inherent element of a culture, we used the social media platform Twitter to compare moral behaviors of Japanese tweets with English tweets. The five basic moral foundations, i.e., Care, Fairness, Ingroup, Authority and Purity, along with the associated emotional valence were compared between English and Japanese tweets. The tweets from Japanese users depicted relatively higher Fairness, Ingroup, and Purity, whereas English tweets expressed more positive emotions for all moral dimensions. Considering moral similarities in connecting users on social media, we quantified homophily concerning different moral dimensions using our proposed method. The moral dimensions Care, Authority and Purity for English and Ingroup, Authority and Purity for Japanese depicted homophily on Twitter. Overall, our study uncovers the underlying cultural differences with respect to moral behavior in English- and Japanese-speaking users.
\end{abstract}


\keywords{morality \and MFD \and J-MFD \and vader \and oseti \and moral emotions \and moral homophily \and culture}

\section{Introduction}
Morality is a term that refers to principles that act as guiding factors in the process of making social judgements, i.e., deciding what is right or what is wrong. Every individual has moral values, which plays a key role in developing human behavior \citep{lagerspetz2016evolution}. Morality has been suggested \citep{haidt2004intuitive, haidt2007morality} as a combination of five basic moral foundations: Care (or Harm), Fairness (or Cheating), Ingroup (or Outgroup), Authority (or Subversion) and Purity (or Degradation). In general, Care pertains to the feeling of protecting the vulnerable and Fairness to doing the right thing, Ingroup relates to exhibiting loyalty to a social group; Authority refers to respecting and obeying tradition; and Purity is the feeling of antipathy toward disgusting matters. The degree of endorsement to these five dimensions may vary from one individual to another \citep{lifton1985individual}, which has often been linked with cultural diversity \citep{jia2017recognizing}. This cultural diversity enables the existence of humans with different beliefs and values \citep{alsheddi2020does}. 

Several researchers have focused on studying morality across different cultures based on moral dilemmas (following one moral value results in violating another,~\citet{awad2018moral}), moral identity (level of importance of moral values to a person's identity, \citet{alsheddi2020does}), and moral concerns (level of concern towards being right or wrong,~\citet{vasquez2001cultural}). A large online experiment was conducted \citep{awad2018moral} across more than 200 countries to capture the essential moral preferences, to be used in the design of new-age machines, such as autonomous vehicles and self-driving cars. The experiments indicated the morality-based cultural differences, as the responses from all the countries could be divided into three different cultures (or clusters); Western (e.g., the United States), Eastern (e.g., Japan) and Southern (e.g., France). \citet{alsheddi2020does} compared the moral identity of two diverse cultures, Islamic and English, specifically between people from Saudi Arabia and Britain. The two moral attributes of care and justice were found to be equally important for participants of both countries. However, the meaning and scope of morality may vary across different cultures \citep{graham2011mapping}. These differences in moral concerns have been studied concerning Americans, representing Western culture, and Filipinos, representing non-Western culture \citep{vasquez2001cultural}. It was found that Americans focused more on Fairness, whereas Filipinos focused on all moral attributes.

Morality has been studied from various perspectives, including emotions in moral context~\citep{brady2017emotion} and moral homophily \citep{dehghani2016purity}. One of the key concepts is the association between morality and emotions~\citep{moll2002neural}. Emotions in psychology \citep{apa_emotion2021} have been defined as a complex phenomenon involving three key components: behavioral, physiological and empirical. All these components combined in the form of emotions are the major drivers in developing moral values towards social events \citep{moll2002neural}. The moral behavior of an individual is associated with various negative as well as positive emotions \citep{tangney2007moral}. Emotions also control one's moral judgement towards various political and social issues, such as gun violence, immigration rights, and governmental policy support~\citep{horberg2011emotions}. In an analysis of morally loaded Americans tweets, it was found that emotions also play a vital role in their spread throughout the social network \citep{brady2017emotion}. On the contrary, the motivation to spread emotional content on social media platforms has been attributed to bonding among people with similar moral characteristics, i.e., moral homophily~\citep{vaisey2010can}. Although sociologists have targeted moral homophily, in general, \citep{mccroskey1975development,vaisey2010can,ing2019offline}, there is a lack of research analyzing homophily concerning different moral attributes. One study found Purity as the only moral attribute that seems to play a key role in connecting individuals, both experimentally and in data from Twitter~\citep{dehghani2016purity}. To measure homophily, moral differences between individuals were compared with their social distance based on their follower-followee relationships on Twitter.

The majority of the literature on moral psychology has focused on participants from Western countries, especially those belonging to the WEIRD society \citep{henrich2010most}. Here WEIRD corresponds to the section of society consisting of people who are Western, educated, industrialized, rich and Democratic. However, few researchers have studied morality in the context of Asian cultures \citep{bespalov2017life, kitamura2021development,matsuo2021appraisal}. Considering language as a proxy for culture \citep{kramsch2011language}, we focused on comparing the moral concerns of Japanese- and English-speaking individuals. In this regard, we used the Twitter platform and collected both Japanese and English tweets related to moral discussions. Apart from comparing moral concerns in both English and Japanese language tweets, we also focused on the comparison of two key aspects of morality between the English and Japanese cultures: moral emotions and moral homophily. Thus, we investigated the following:

\begin{enumerate}
    \item Comparison of moral loadings for English and Japanese texts.   
    \item Comparison of emotional valence with respect to each moral dimension.
    \item Measuring homophily for basic moral attributes.
\end{enumerate}

\begin{figure*}
\centering
\includegraphics[width=\textwidth]{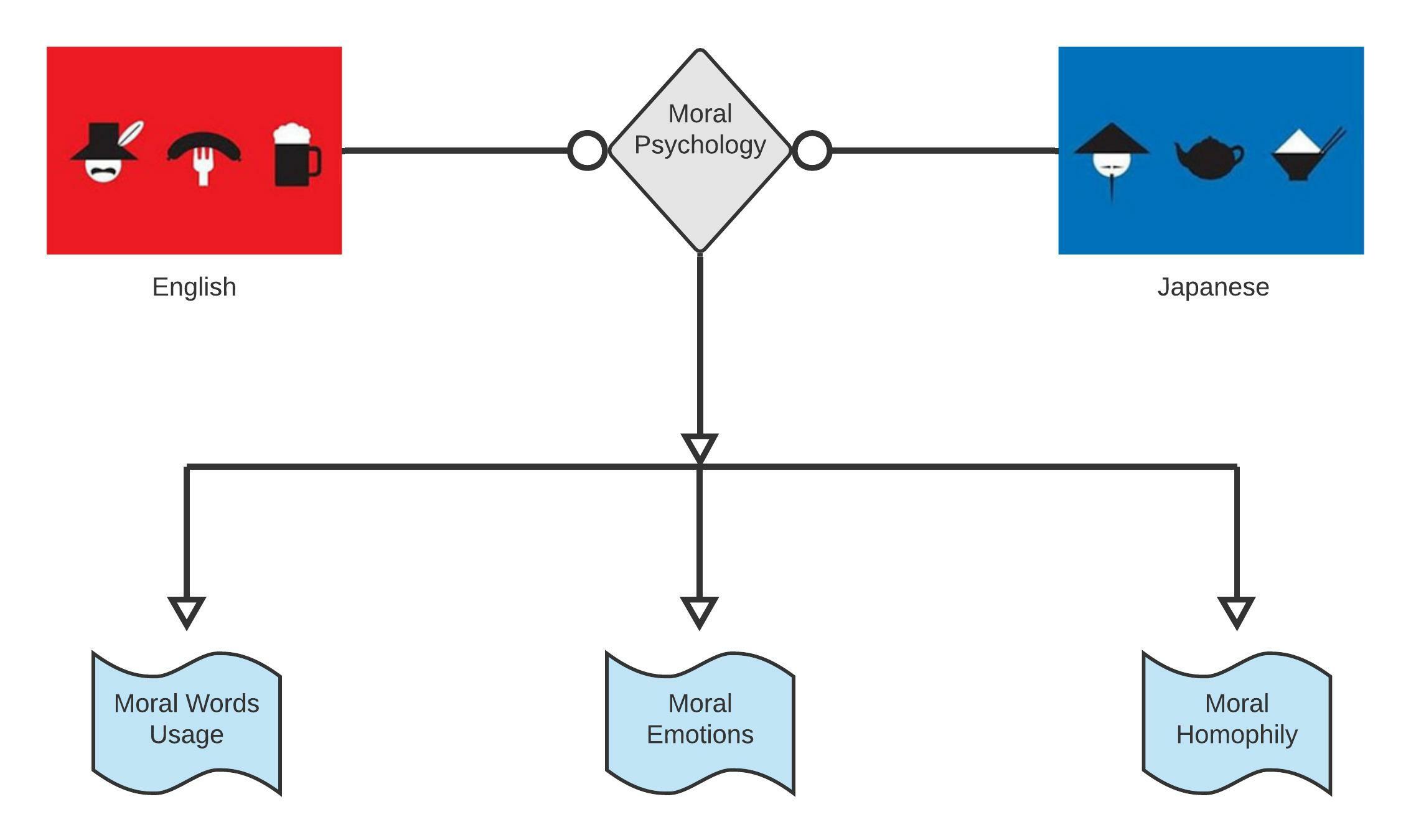}
\caption{Comparison of Psychological Aspects of Morality between English and Japanese culture.} \label{fig:contri}
\end{figure*}

This research is significant in the field of moral psychology, as it studies multiple psychological aspects of morality (Figure \ref{fig:contri}). These aspects are analyzed and compared for English and Japanese cultures represented through the use of language on Twitter.

\section{Materials and Methods}
First, we explain our data collection method, followed by the procedure used for identifying moral foundations. Next, we describe our emotion extraction process for both English and Japanese tweets. Finally, we specify the measures for computing homophily related to the five basic moral foundations with the help of a retweet network.

\subsection{\textbf{Data}}
As the aim of the study is to compare the moral foundations of the English and Japanese languages, we used the social media platform Twitter to extract English and Japanese posts (tweets). We collected tweets that express moral concerns with words related to morality. For English tweets, keywords such as \say{moral}, \say{immoral}, \say{morality} and \say{immorality} were used. The meaning words and their synonyms in Japanese like \begin{CJK}{UTF8}{min}
\say{道徳}, \say{背徳}, \say{非道徳}, \say{反道徳}, \say{不道徳}\end{CJK} were used for extracting Japanese tweets. This resulted in 1,006,886 English and 745,673 Japanese tweets in total, collected over a period of approximately six months i.e from March 1, 2016 to September 24, 2016. For further analysis, all the collected tweets were pre-processed by removing URLs, mentions, hashtag symbols, stop words and any unwanted symbols. 

\subsection{Dictionaries: MFD and J-MFD}
We used two moral foundations dictionaries to quantify moral loadings from the text. The first is the original Moral Foundation Dictionary (hereafter MFD) and the second is the Japanese version of the MFD (hereafter J-MFD). The original MFD \citep{haidt2007morality} was developed for the English language, comprising 156 words (e.g., kill) and 168 word stems (e.g., killer*) pertaining to virtues and vices of each of the five moral dimensions along with other general morality related terms ({\small \url{https://moralfoundations.org/wp-content/uploads/files/downloads/moral\%20foundations\%20dictionary.dic}}). The J-MFD \citep{matsuo2019development} was developed using all the moral terms (i.e., words and word stems) from the original MFD.
The J-MFD consists of 718 moral terms in total ({\small \url{https://github.com/soramame0518/j-mfd}}). Both the MFD and J-MFD were used to study moral foundations from the text  \citep{haidt2007morality,dehghani2016purity,kaur2016quantifying,matsuo2021appraisal}. 

Although we are aware of other variations of the MFD \citep{garten2016morality,araque2020moralstrength,hopp2021extended} for English, the same is not available for Japanese. 
In addition, the J-MFD has been well validated~\citep{matsuo2019development} and applied to analyze Japanese tweets~\citep{matsuo2021appraisal}.
Therefore, we decided to use the original MFD and its counterpart, the J-MFD, to ensure equivalence in morality measurement methods.

\subsection{Moral Loadings of Tweets}\label{sec:moralload}
The overall moral loadings of all the tweets were extracted. For this, we were concerned with the five basic moral foundations (dimensions): Care, Fairness, Ingroup, Authority, and Purity. For a tweet $t_i$, the moral loading ($ml_{ij}$) with respect to a moral dimension $d_j$ is given as:
\begin{equation}
\label{eq:moral_loading}
    ml_{ij} = \frac{|W_{d_j}(t_i)|}{|W(t_i)|}
\end{equation}
where $W_{d_j(t_i)}$ is a set of words from a dictionary occurring in tweet $t_i$, corresponding to dimension $d_j$ and $W(t_i)$ is the set of words in tweet $t_i$, corresponding to all five moral foundations. In Eq.~\ref{eq:moral_loading}, both virtue and vice words from the dictionary were combined for each moral category. To find out which moral dimension(s) was represented by the given tweet $t_i$, the one with the highest value of $m_{ij}$ was selected. If the given tweet had more than one dimension with the same scores, then the given tweet was loaded with multiple moral dimensions. 

The above approach was used for both English and Japanese tweets. If any given tweet did not contain any of the words from the dictionary belonging to the five basic moral dimensions, we simply filtered such tweets from our database. This filtering resulted in 364,053 English tweets and 321,290 Japanese tweets. The number of users in English tweets were 254,422 (with 98\% users tweeted/retweeted $\leq$ 5 times), whereas in Japanese tweets there were 149,529 users (with 95\% users tweeted/retweeted $\leq$ 5 times). Since, our study aims to highlight the morality based on differences between Western and non-Western (Japanese in our case) cultures, it is vital to ensure that the users of English tweets must be representative of Western culture. This is because, while Japanese tweets may be mainly posted by Japanese users, the same may not be true for English tweets, as English tweets may be posted by users belonging to different countries. Therefore, we randomly extracted the location of 10,000 users, i.e., approximately 5\%, based on the approach used in \citep{singh2020multidimensional}. It was found that more than 80\% of the users of English tweets were from Western countries \citep{western}, of which the majority were from the US (Figure S1). This was expected as the US ranks first in the world in the number of users on Twitter \citep{activeusers}. Thus, we can say that the users of English tweets are good candidates for representing Western culture.

\subsection{Quantifying Emotions in Morally Loaded Tweets}
\label{sec:emotions}
Considering emotion to be a binary entity, i.e., positive or negative, it would be interesting to see which moral dimensions are expressed generally with positive emotions and, similarly, which dimensions are conveyed with with negative emotions. For English tweets, we used the VADER sentiment analysis tool \citep{hutto2014vader}. The VADER model uses a dictionary containing words along with their emotional intensity and applies grammatical and syntactical rules to detect binary emotion in texts. In the case of Japanese tweets, \textit{oseti}, a dictionary-based tool specifically developed for the Japanese language, was used. The tool uses a Japanese polar dictionary \citep{kobayashi2004collecting,higashiyama2008learning} for detecting emotional valence. Both methods return scores in the range of -1 to +1; -1 indicates that the given tweet expresses extremely negative emotions, whereas +1 indicates extremely positive emotions.

\subsection{Moral Homophily on Online Social Networks}
\label{sec:homophily}
The degree of homophily for the five moral foundations was measured individually. In this regard, we computed the moral loadings, with respect to all five moral dimensions, for each user based on their tweets. The complete procedure is also given in Algorithm \ref{alg:proportion}. First, we identified users in our dataset with more than one moral tweet (i.e., tweet labelled with one or more moral dimensions in Section \ref{sec:moralload}). We followed this procedure for the English and Japanese tweets, separately. Then, we segregated those tweets for each user based on the labelled moral dimension. The loading of each moral dimension $j$ for a given user $i$ is computed as a share of tweets representing moral dimension $j$ for user $i$ with reference to the overall number of morally labelled tweets by the given user.

\begin{algorithm}
	\caption{Computing Moral Loadings of a User} 
	 \begin{flushleft}
    \textbf{Input} Set of morally labelled tweets\\
    
    \textbf{Output} Overall loading of five moral dimensions of authors of morally labelled tweets 
    \end{flushleft}
	\begin{algorithmic}[1]
		
		\State Let $T$ be the set of morally labelled tweets.
		\State Let $k$ be the unique number of authors of all the tweets in $T$.
		\State Divide $T$ into $k$ disjoint set of tweets  $\langle Tu_1$,$Tu_2$,...,$Tu_k\rangle$ such that each set $Tu_i$ consists of morally labelled tweets from user $i$. 
        \State Further, divide each set $Tu_i$ into five subsets such that each subset $Tu_{ij}$ corresponds to the set of morally loaded tweets from user $i$ representing moral dimension $j$.
        \State The loading of moral dimension $j$ in user $i$ is shown in the following equation:
            \begin{equation}
                mp_{ij} = \frac{|Tu_{ij}|}{|Tu_i|}
            \end{equation}
        \State After combining the moral loadings computed for each dimension in the previous step, the moral loading of a user $i$, $mp_i$ will be represented as:-

        \begin{equation}
            mp_i=<mp_{i1}, mp_{i2},mp_{i3},mp_{i4},mp_{i5}>
        \end{equation}

	\end{algorithmic}
	\label{alg:proportion}
\end{algorithm}

Next, using the moral loading of each user computed above, we find the moral dimension with the maximum value for each user. Here, we have assumed that the moral dimension that best represents the given user is the one with the maximum proportion in their tweets and hence the user is labelled accordingly. If, for any user, this results in more than one moral dimension, due to equal proportions, then those users were excluded from the study. To avoid noisy data and ensure quality labelling, users with only single morally labelled tweet were also filtered from our analysis. We then constructed a retweet network, where each user is represented by nodes and an edge between any two users denotes that one of them retweeted the tweet posted by the other. The retweet network for English tweets comprised 5,107 nodes and 10,241 edges, whereas for Japanese tweets, the network consisted of 6,329 nodes and 19,893 edges. The moral homophily on a retweet network, concerning each moral foundation was computed using Algorithm \ref{alg:homophily}. We first generated the set of edges for each user, that connects them to another user with similar moral labels (Eq. \ref{eq:edge}). Based on the weights of such edges, the homophily for each user was computed (Eq. \ref{eq:node_homo}). Finally, the network homophily for a given moral foundation $j$ was then measured by averaging the homophily scores for each user, labeled with moral foundation $j$ (Eq. \ref{eq:homo}).   

\begin{algorithm}
	\caption{Quantifying Moral Homophily} 
	 \begin{flushleft}
    \textbf{Input} A retweet network G (or any other social network)
    
    \textbf{Output} Homophily scores for each moral foundation 
    \end{flushleft}
	\begin{algorithmic}[1]
		
		\State Let $N$ and $E$ be the set of nodes and edges for a given network $G$.
		\State Divide E into $|N|$ non-disjoint sets $<E_1,E_2,.,E_i,.,E_{|N|}>$ such that $E_i$ is the set of edges from node $i$ to other nodes in the network $G$. 
        \State For each node $i$, we compute $E_i^D$ as follows:-
        \begin{equation}\label{eq:edge}
            E_i^D = \{e(i,j) \, | \, e(i,j) \in E_i, D_i = D_j \, \text{and} \, i\neq j\}
        \end{equation} 
        where the moral label of node $i$, $D_i=\argmaxA_i mp_i$ and $e(i,j)$ is an edge between node $i$ and node $j$.
        \State For each node $i$, we computed its homophily using the following equation:
        \begin{equation}\label{eq:node_homo}
            h_i = \frac{\sum_{j=1}^{|E_i^D|} w_{ij}}{\sum_{j=1}^{|E_i|} w_{ij}}
        \end{equation}
        where $w_{ij}$ is the weight of the edge $e(i,j)$. 
        \State The overall score of homophily for network $G$ w.r.t. dimension $j$ can be computed as:
        \begin{equation}\label{eq:homo}
            H_j=\frac{1}{|N_j|}\sum_{i=1}^{|N_j|} h_i
        \end{equation}
        where $N_j=\{n\,| \, n \in N \, and \, D_n=j\}$.
		
	\end{algorithmic}
	\label{alg:homophily}
\end{algorithm}

\section{Results}
\subsection{Comparing Moral Foundations}
\label{sec:moral_found}
\begin{figure*}
\centering
\includegraphics[width=\textwidth]{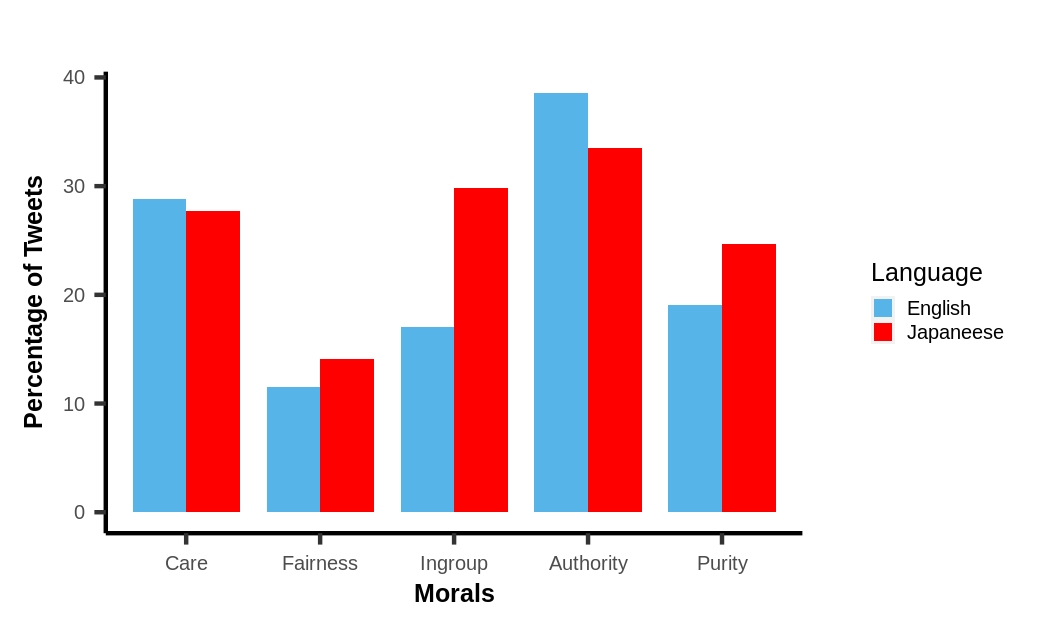}
\caption{Moral Foundations of English and Japanese Tweets.} \label{fig:moral}
\end{figure*}

The comparison of the five moral dimensions with respect to English and Japanese is shown in Figure \ref{fig:moral}. For both languages, the maximum number of tweets represented ``Authority'', whereas ``Fairness'' was represented the least. However, if we compare each moral dimension, the language used in the English tweets represented relatively higher moral attributes, such as Authority and Care, whereas attributes like Ingroup, Fairness and Purity were represented more in Japanese tweets. We, therefore, conducted a statistical comparison between the English and Japanese tweets to identify the underlying differences between each moral attribute. As the moral loadings obtained from Section \ref{sec:moralload} do not follow normal distribution for each dimension, we performed the Kruskal-Wallis test \citep{daniel1990kruskal}, a non-parametric significant test. The loadings of all five moral foundations ($\chi^{2}$: 26.23 (Care), 3571.30 (Fairness), 23863.00 (Ingroup), 2605.60 (Authority) and 5691.50 (Purity); $p < 0.001$) varied significantly for English and Japanese tweets, indicating the moral difference between the cultures of Japanese and English speaking users. The higher value of statistics in the test indicates a greater difference in the loadings of corresponding moral dimensions. Thus, out of all five dimensions, the maximum difference between the moral concerns of English and Japanese people was for Purity. This could be due to the different interpretations of pure and impure among Japanese people as compared to Western culture \citep{kitamura2021development}. 

\begin{figure*}
\centering
\includegraphics[width=\textwidth]{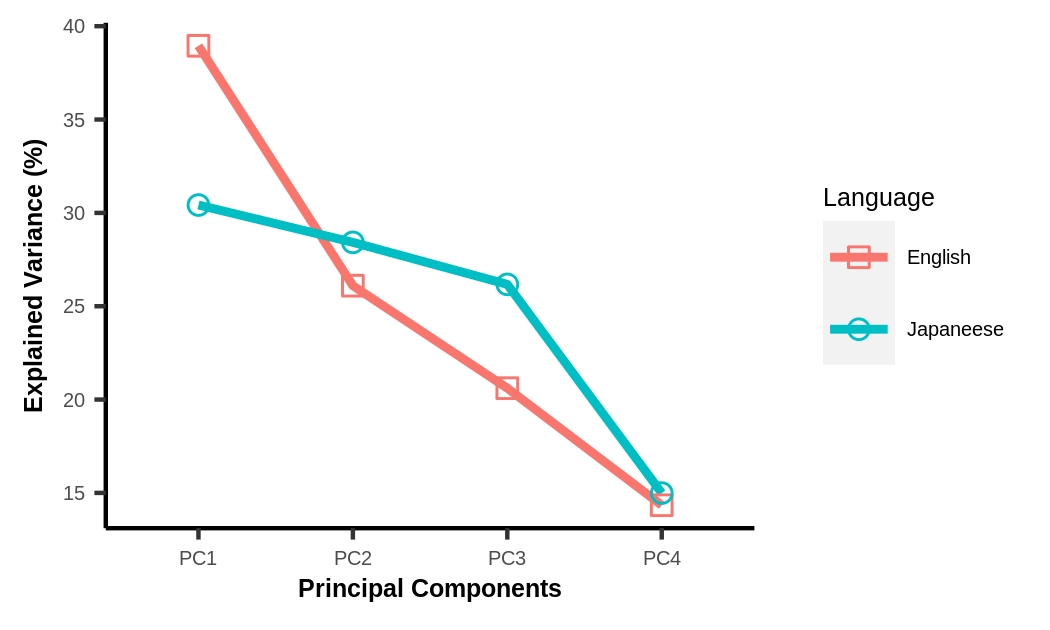}
\caption{Scree Plot for Principal Components of moral loadings of English and Japanese tweets.} \label{fig:scree}
\end{figure*}

In our study, morality was represented as a five-dimensional vector, in which each vector component denoted one of the five moral foundations. To visualize the correlations among moral foundations, we employed the Principal Component Analysis to transform \citep{wold1987principal} the given moral loadings. The transformed space for both English and Japanese tweets resulted in four principle components (hereafter PCs), explaining the overall variance in the five moral dimensions (Figure \ref{fig:scree}). As seen in Figure \ref{fig:scree}, the first two PCs for English tweets can explain 65.8\% of the variance and similarly for Japanese tweets, the first two PCs explain 58.6\% of the variance.

\begin{figure*}
\centering
\includegraphics[width=\textwidth]{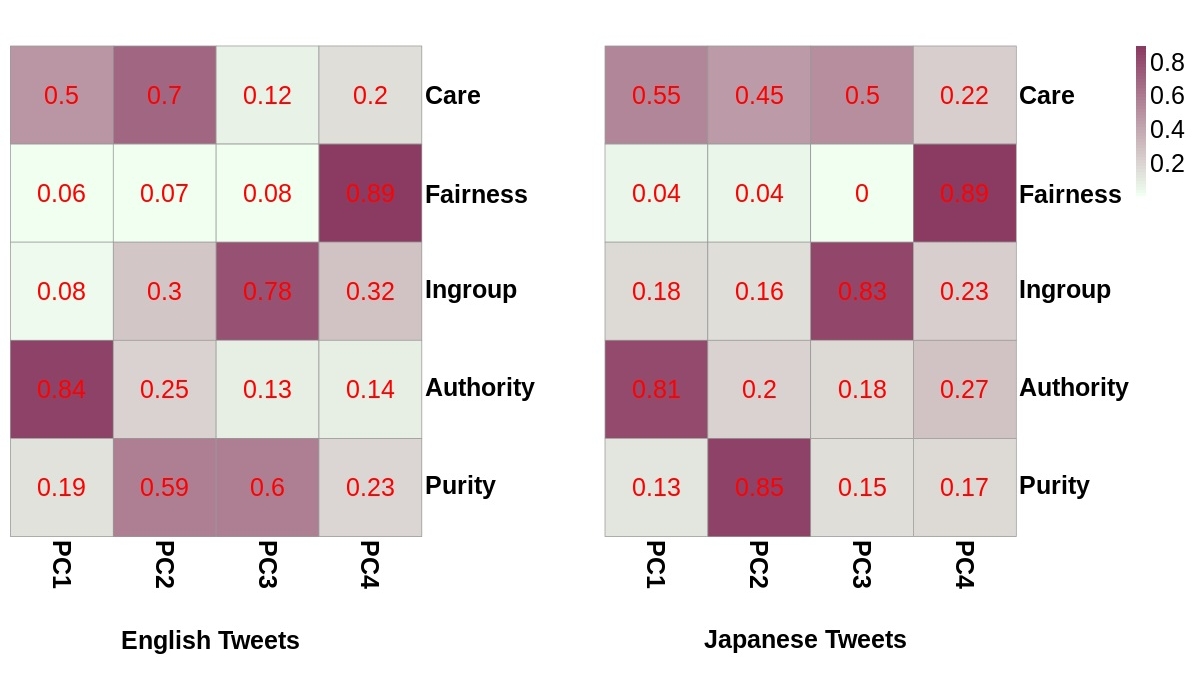}
\caption{Contribution of moral dimensions to each principal component} \label{fig:heatmap}
\end{figure*}

Next, we looked at the contributions of the moral dimensions in these PCs. For this, we generated a heatmap between the five moral foundations and all four PCs for both the English as well as Japanese tweets (Figure \ref{fig:heatmap}). The values in each cell of the heatmap were kept absolute, as they will clearly depict the major and minor contributions of all moral dimensions. As observed for both heatmaps, Authority, Fairness, and Ingroup mainly contributed to PC1, PC4 and PC3, respectively. However, for the Japanese tweets, the Purity dimension was mainly explained by PC2, whereas the Purity dimension in the English tweets was explained by both PC2 and PC3. In regard to the Care dimension, both PC1 and PC2 were required to explain the variance in the case of the English tweets. The Care dimension in the Japanese tweets contributed to PC1, PC2 and PC3. 

\begin{figure*}
\centering
\includegraphics[width=\textwidth]{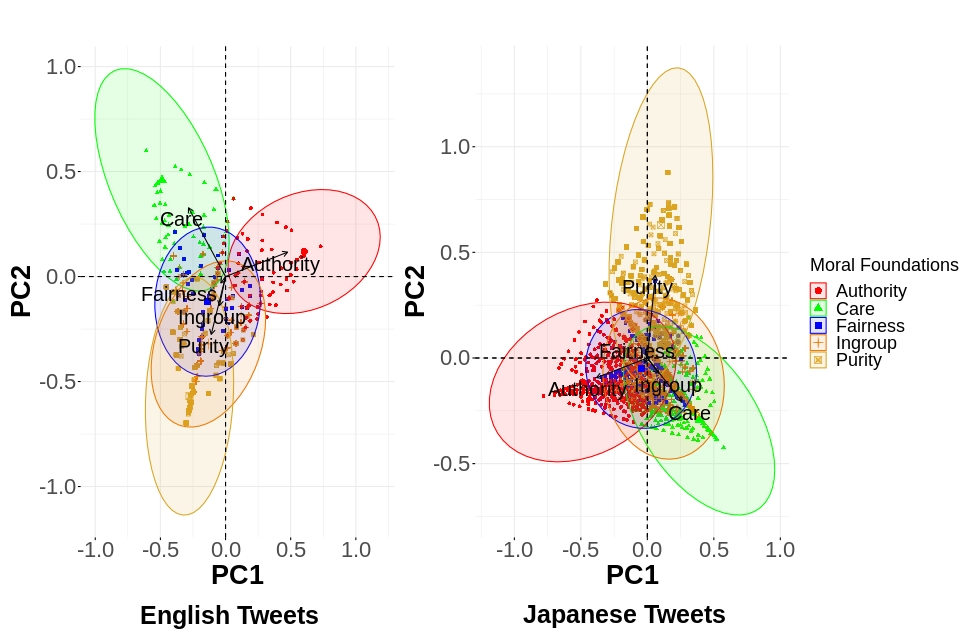}
\caption{PCA Biplots with PC1 and PC2 for English and Japanese Moral Loadings.} \label{fig:biplot12}
\end{figure*}

The heatmaps generated above elucidate the contributions of all the moral dimensions into the orthogonal vectors of the transformed space. To establish the relation between each dimension through these principal components, we created biplots, which are two-way plots displaying both the scores of the PCs as well as the loading vectors into one single plot. The PCA biplots for the English and Japanese tweets between the first and second principal components are shown in Figure \ref{fig:biplot12}. In the English tweets, Ingroup and Purity seem to be in positive correlation with each other. This might be because that Purity has been found to be one of the factors that divides people \citep{dehghani2016purity}. Thus, these two attributes are generally triggered simultaneously for morality-based discussions among users of English tweets. We did not observe any other correlations for moral attributes in the English tweets. After analyzing the biplot for the Japanese tweets, it was observed that Care and Ingroup were positively correlated with each other. Japanese culture is famous for loyalty (or Ingroup) \citep{graham2011mapping}, but at the same time, death from overwork, i.e., karoshi, is one of the social problems faced among the Japanese \citep{north2016hope}. Therefore, it might be possible that people have shown care/harm along with loyalty in their discussions. Thus, moral foundations ($d=5$) in both English and Japanese tweets can be represented with lower dimensions ($d=4$). This may be because that concern for Ingroup is generally accompanied with concern for other moral dimensions, i.e., Purity for English tweets and Care for Japanese tweets. Although the first two PCs cover only around 60\% for both English and Japanese tweets (for the biplots between other PCs please refer Figures S2-S6), these observations can act as building blocks for further research in this direction. Therefore, the different patterns observed in the moral dimension reduction also suggest the existence of cultural differences between Japanese- and English-speaking users. 

\subsection{Moral Emotions}
\begin{figure*}
\centering
\includegraphics[width=\textwidth]{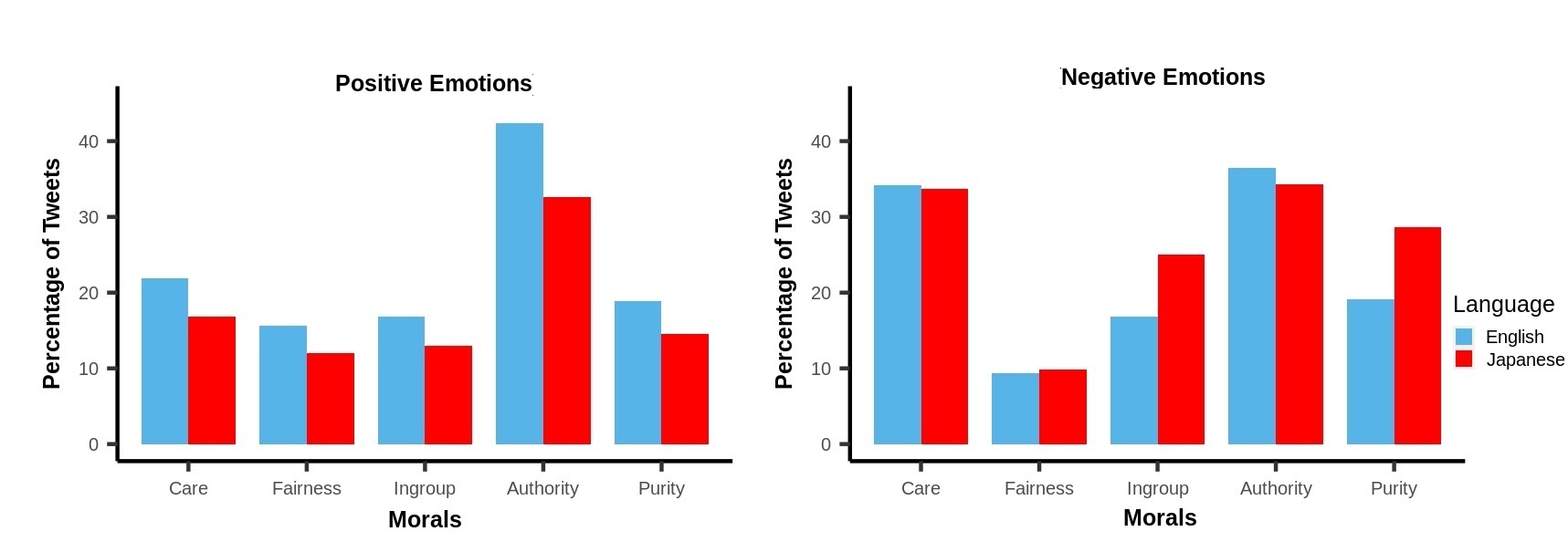}
\caption{Distribution of tweets with positive emotions (left) and tweets with negative emotions with respect to five moral dimensions.} \label{fig:emotion}
\end{figure*}

The moral discussions on an online platform related to different topics highlight the emotions of users engaged in those discussions. 
Next, we have analyzed the emotional valence of tweets related to morality. Using the intensity of binary emotions (i.e., positive or negative) obtained from Section \ref{sec:emotions}, we labelled all the tweets as either positive, negative or neutral. If the emotional valence of the tweet was below 0, we labelled it as negative and if it was above 0, the tweet was considered as positive. All the remaining tweets were considered to be neutral in expressing emotions \citep{hutto2014vader,higashiyama2008learning}. After comparing the percentage of positively labelled tweets for each moral foundation between English and Japanese, Figure \ref{fig:emotion} shows that English tweets expressed positive emotions than Japanese tweets for each moral dimension. However, if we compare tweets expressing negative emotions, Japanese tweets had a relatively higher share of tweets representing moral attributes like Purity and Ingroup, in comparison to English tweets. To find statistically significant emotional differences in discussing morality between users of English and Japanese tweets, we performed the Kruskal-Wallis test on the emotional valence of tweets for the five moral dimensions. The test statistic values obtained for all five foundations ($\chi^{2}$: 21477.00 (Care), 4060.80 (Fairness), 20685.00 (Ingroup), 9520.80 (Authority) and 6737.50 (Purity); $p < 0.001$) signify the differences between Japanese- and English-speaking users in expressing emotions on Twitter. 

\subsection{Moral Homophily}
\begin{figure*}
\centering
\includegraphics[height=20cm,width=14cm]{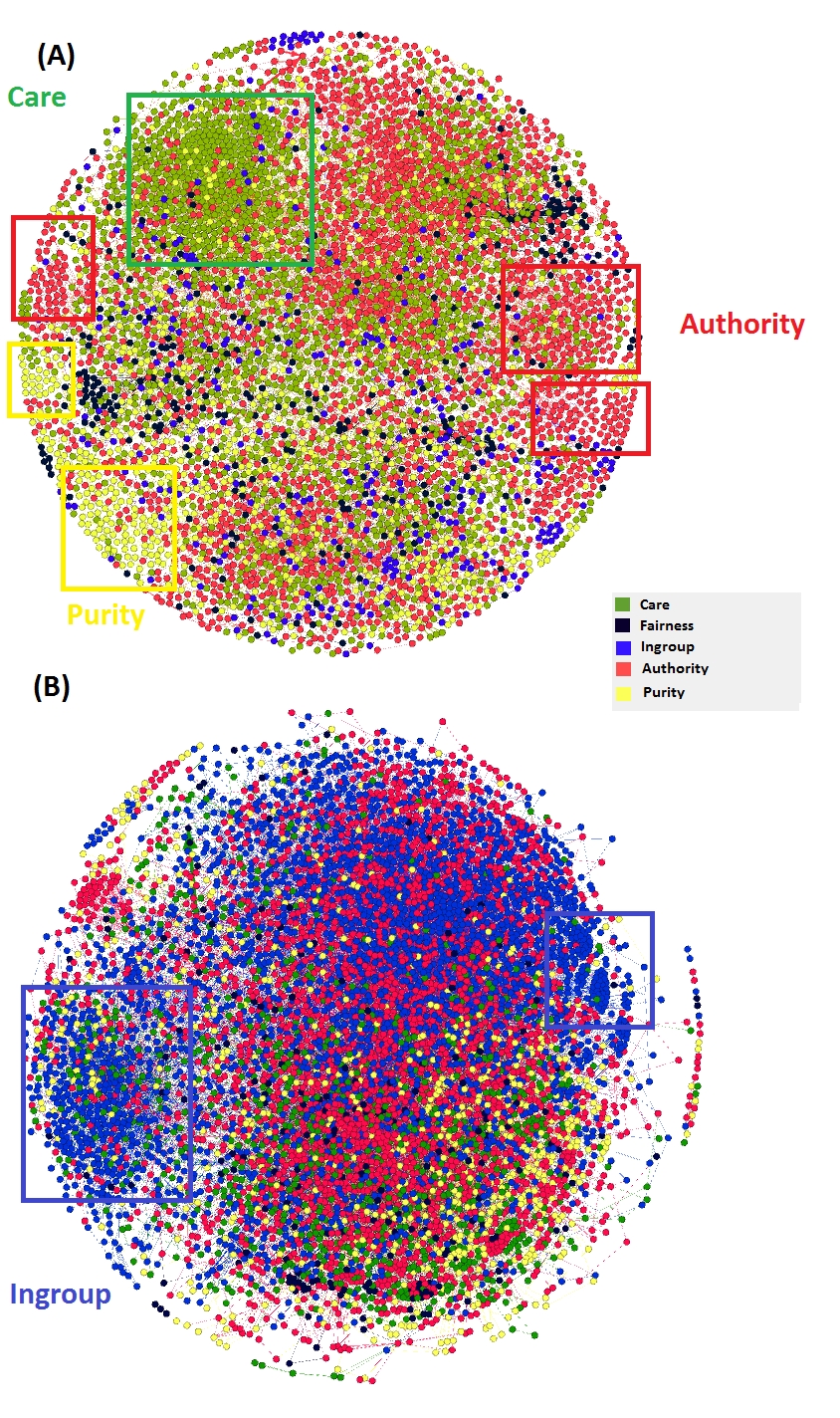}
\caption{Retweet Network (2-core) for (a) English and (b) Japanese tweets. Some rectangular boxes are also drawn to highlight the clustering of similar colored nodes along with their corresponding moral label.} \label{fig:network}
\end{figure*}

All individuals have a tendency to be attracted to those with whom they share similar characteristics \citep{bollen2011happiness}. 
As shown in previous research \citep{mccroskey1975development}, it is possible that morality could be one of the factors that leads to a homophilous social environment. We examined the existence of homophily concerning all five moral dimensions in English and Japanese. 
For this, we choose a retweet network, as it clusters users who share similar characteristics or interests via. retweets on Twitter. Previous research has mainly focused on the ideology aspect behind the community structures of such networks. Our focus, however, is on the moral aspects. The level of homophily for each moral foundation is computed using Algorithm \ref{alg:homophily}. The moral dimensions Authority and Purity seems to be homophilous (scoring greater than 0.5) for both English and Japanese speaking users. Furthermore, Care in English tweets and Purity in Japanese were found to attract users on Twitter. The same observations can also be visualized from their respective retweet networks (Figure \ref{fig:network}). For simplicity, we have displayed a two-core network \citep{bader2003automated} using Gephi software \citep{bastian2009gephi}, with the Frutcherman Reingold layout. 

\begin{table}[H]

    \caption{Homophily Scores for five basic moral foundations.}
    \label{tab:homophily}
    \centering
    \begin{tabular}{cccccc}
    \hline
    \textbf{} & \textbf{Care} & \textbf{Fairness} & \textbf{Ingroup} & \textbf{Authority} & \textbf{Purity}\\
    \hline
    \textbf{English} & \textbf{0.68} & 0.49 & 0.34 & \textbf{0.68} & \textbf{0.59}\\
    \textbf{Japanese} & 0.29 & 0.40 & \textbf{0.69} & \textbf{0.65} & \textbf{0.60}\\
    \hline
    \end{tabular}
\end{table}

\section{Discussion}
Our society consists of diverse cultures with their own ideas and beliefs. Hence, moral appraisals may vary from one culture to another. In this study, considering language as a prototype of culture, we compared the moral behavior of two different cultures: Western culture represented through English tweets and non-Western culture represented through Japanese tweets. Our findings suggests that English users were relatively more concerned for Care and Authority than Japanese users. Conversely, the Japanese were relatively more concerned with Fairness, Ingroup and Purity compared to English users. A higher inclination of Eastern culture towards Ingroup and Purity, in comparison to Western, has been previously found \citep{graham2011mapping}. Although Japanese is one of the cultures from the East, these findings may highlight changing moral concerns among the Japanese \citep{hamamura2012cultures}. Further research in this direction is required to observe moral differences within multiple cultures of Eastern countries. 

We also assessed the emotions in the tweets of both languages and found that the users of the Japanese tweeters used relatively more negative emotion-based words than the users of English tweets. This could be due to the desire of Japanese people to express fewer positive emotions in general \citep{safdar2009variations}, although there has been lack of research on emotions in a moral context using online conversation \citep{brady2017emotion,brady2020mad}. 

We also detected homophily in the English and Japanese retweet networks with respect to the five moral foundations. The results highlighted the existence of homophily among English users for Care, Authority and Purity. Previous work~\citep{dehghani2016purity} has observed the Purity homophily (i.e., Purity as a morality for social biding) using English tweets. The differences in the observations could be due to two reasons: (1) Our study focuses on retweet networks instead of followers network, as retweeting behavior is more polarizing in nature \citep{conover2011political,Sasahara2020}, and (2) The difference in the data itself, as our data is more generic and consists of tweets discussing morality. Similar to users of English tweets, Japanese users also depicted homophily for Authority and Purity. Besides these dimensions, Ingroup was also found to be homophilous among the Japanese, which may be because of the significance of loyalty in their culture \citep{ernayani2021loyalty}. In this study, we have used a weighted version of a retweet network to compute the homophily score, as our dataset is longitudinal and covers approximately six months. In a small dataset (e.g., tweets that cover a few days), accidental events (typhoon, election, car crash, etc.) may cause an overwhelming amount of retweets, which are not necessarily related to homophily. In such a case, a non-weighted version of a retweet network could be a better alternative for the homophily score.  

Overall, our results indicate the moral differences between the non-Western and Western cultures. It is noted that non-Western culture is analyzed through Japanese tweets, most likely posted by users from Japan. To the contrary, Western culture is analyzed through English tweets, largely posted by the users from the US. However, cross-cultural differences in moral behaviors, from different outlooks, is depicted through our findings. We expect our work will motivate other researchers of moral psychology to focus on more studies that include non-WEIRD participants and encourage them to use data from social media platforms to assess various cross-cultural social and psychological phenomena at scale.

\section{Conclusion}
This article analyzes the multiple aspects of moral psychology to compare the English and Japanese cultures. To this end, moral conversations in both languages were collected from the Twitter platform. Based on the frequency of moral words from the MFD and J-MFD, it was observed that Japanese users are more concerned with the moral foundations of Fairness, Ingroup, and Purity than the English users. Moreover, the English tweeters have relatively positive feelings towards discussing morality on social media. We were also able to detect homophily based on moral foundations in the retweet networks of English and Japanese tweets. The cross-cultural moral differences observed online between English and Japanese cultures can be further consolidated by surveys or questionnaires in future research. 

\section*{Acknowledgement}
This work was supported by JSPS/MEXT KAKENHI (Grant Numbers JP18H01078, JP19H04217, and JP17H06383 in \#4903). 
KS thanks the members of JST CREST project (Grant Number JPMJCR17A4) for discussions.


\bibliographystyle{unsrtnat}
\bibliography{morality}  





\newpage
\renewcommand\thesection{}
\section{SUPPLEMANTARY INFORMATION}

\renewcommand{\thefigure}{S1}
\begin{figure*}[h]
\centering
\includegraphics[width=\textwidth]{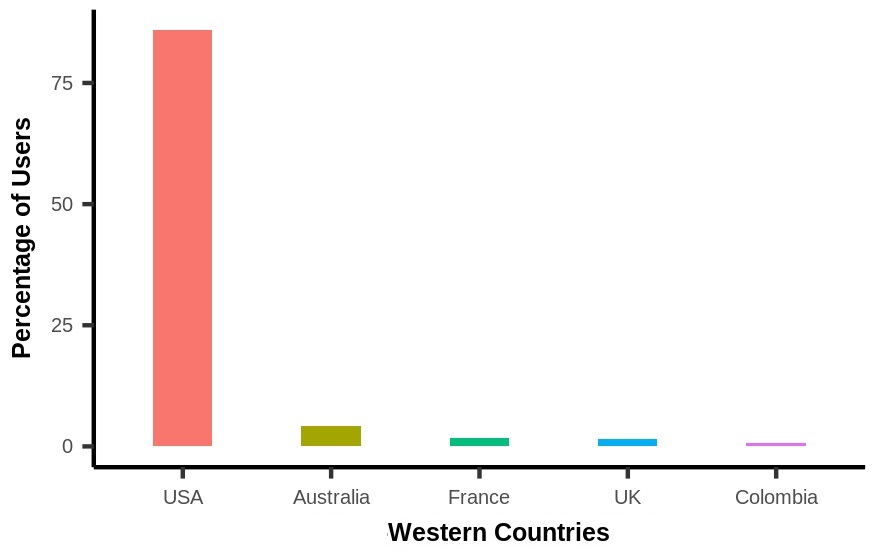}
\caption{Top five countries based on the random selection of users (of English tweets) belonging to western countries.} \label{fig:comb}
\end{figure*}

\renewcommand{\thefigure}{S2}

\begin{figure*}
\centering
\includegraphics[width=\textwidth]{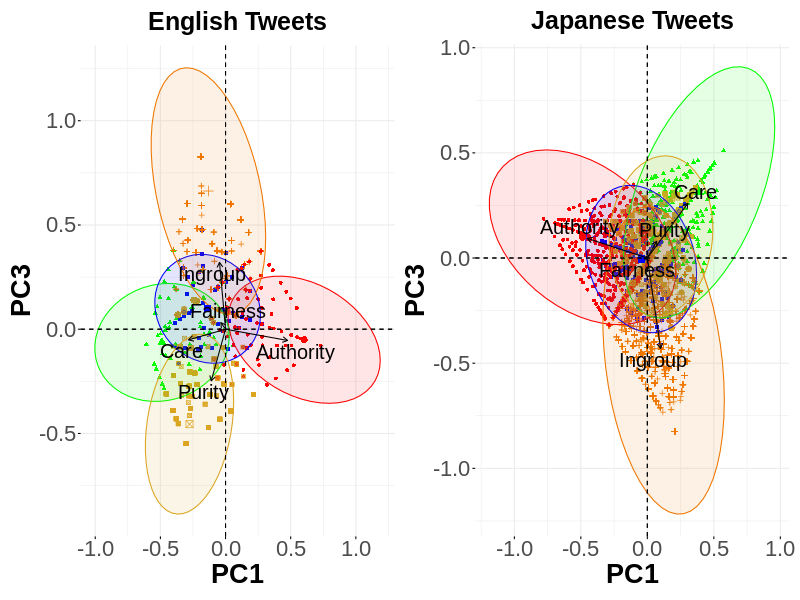}
\caption{PCA Biplot with PC1 and PC3 for English and Japanese Moral Loadings.} \label{fig:biplot13}
\end{figure*}

\renewcommand{\thefigure}{S3}
\begin{figure*}
\centering
\includegraphics[width=\textwidth]{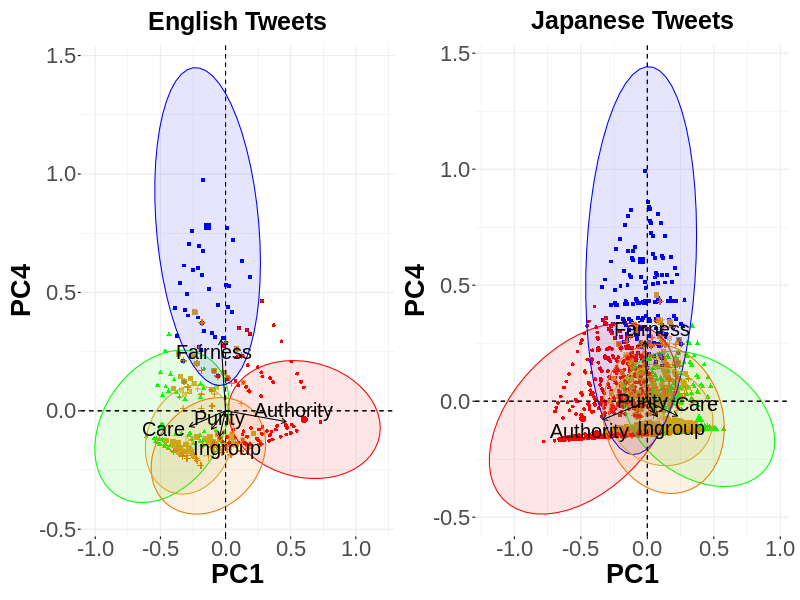}
\caption{PCA Biplot with PC1 and PC4 for English and Japanese Moral Loadings.} \label{fig:biplot14}
\end{figure*}

\renewcommand{\thefigure}{S4}
\begin{figure*}
\centering
\includegraphics[width=\textwidth]{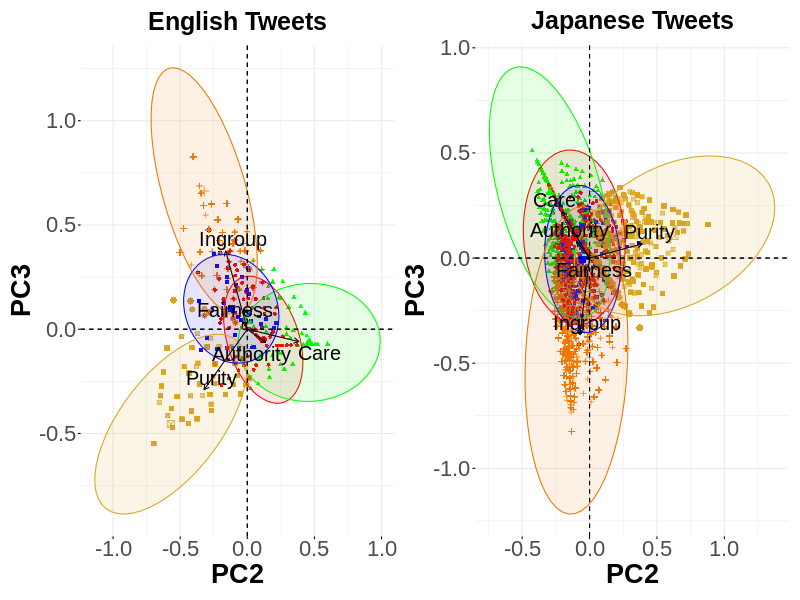}
\caption{PCA Biplot with PC2 and PC3 for English and Japanese Moral Loadings.} \label{fig:biplot23}
\end{figure*}

\renewcommand{\thefigure}{S5}
\begin{figure*}
\centering
\includegraphics[width=\textwidth]{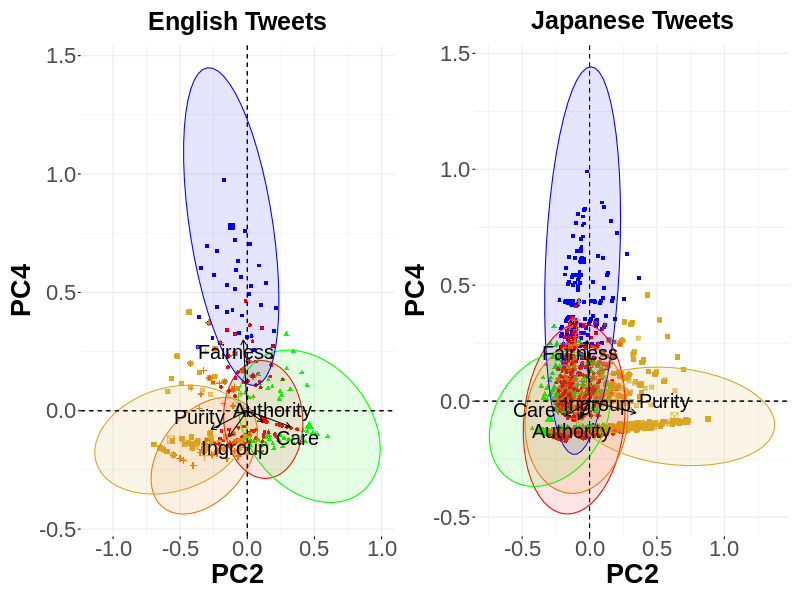}
\caption{PCA Biplot with PC2 and PC4 for English and Japanese Moral Loadings.} \label{fig:biplot24}
\end{figure*}

\renewcommand{\thefigure}{S6}
\begin{figure*}
\centering
\includegraphics[width=\textwidth]{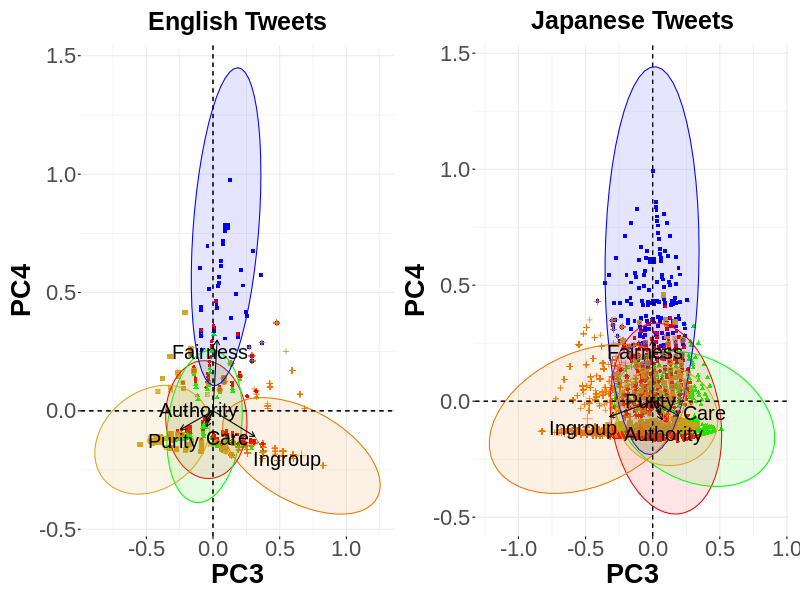}
\caption{PCA Biplot with PC3 and PC4 for English and Japanese Moral Loadings.} \label{fig:biplot34}
\end{figure*}

\end{document}